\documentclass{article}

\usepackage[utf8]{inputenc} %
\usepackage[T1]{fontenc}    %

\usepackage{microtype}
\usepackage{graphicx}
\usepackage{booktabs} %
\usepackage{subcaption}

\usepackage[breaklinks]{hyperref}

\usepackage{iclr2025_conference,times}
\iclrfinalcopy
\renewcommand{\cite}[1]{\citep{#1}}

\usepackage{amsmath}
\usepackage{amssymb}
\usepackage{mathtools}
\usepackage{amsthm}
\usepackage{soul}
\usepackage{xcolor}
\usepackage{nicefrac}       %
\usepackage{microtype}      %
\usepackage[capitalize,noabbrev]{cleveref}

\usepackage{xspace}

\theoremstyle{plain}

\theoremstyle{definition}

\theoremstyle{remark}

\newcommand{\modelname}{MoDE\xspace}
\newcommand{\modelnamefull}{Modular Domain Experts\xspace}

\newcommand{\eg}{e.g.,\xspace}

\newcommand{\myparagraph}[1]{\noindent{\bfseries #1}}

\usepackage{xcolor}
\usepackage{pgfplots}
\pgfplotsset{compat=1.13}
\usepackage{tikz}
\usetikzlibrary{shapes}
\usetikzlibrary{calc}
\usetikzlibrary{trees}
\usetikzlibrary{positioning}
\usetikzlibrary{arrows}
\usetikzlibrary{arrows.spaced}
\usetikzlibrary{arrows.meta}
\usetikzlibrary{chains, matrix}
\usetikzlibrary{shapes, shapes.geometric, shapes.symbols, shapes.multipart}
\usetikzlibrary{decorations.markings, decorations.pathreplacing, decorations.pathmorphing, decorations.text, decorations.shapes}
\usetikzlibrary{backgrounds}
\usetikzlibrary{patterns}
\usetikzlibrary{fit}

\newcommand*{\circled}[1]{
  \protect
  \tikz[baseline=(char.base)]{
    \protect
    \node[
      shape=circle,
      fill=black,
      text=white,
      font=\tiny,
      align=center,
      inner sep=2pt
    ](char){#1};
  }
}

\newcommand{\one}{({\em i}\/)}
\newcommand{\two}{({\em ii}\/)}

\usepackage[textsize=tiny]{todonotes}

\usepackage[shortlabels]{enumitem}
\setlist[enumerate]{itemsep=1.5pt, topsep=2pt, parsep=0pt}
\setlist[itemize]{itemsep=1.5pt, topsep=2pt, parsep=0pt}

\title{Scalable Multi-Domain Adaptation of Language Models using Modular
Experts}

\author{
  \begin{tabular}{lll}
    Peter Schafhalter\textsuperscript{*,\dag,1} &
    Shun Liao\textsuperscript{*,2} &
    Yanqi Zhou\textsuperscript{\ddag,2} \\
    Chih-Kuan Yeh\textsuperscript{2} &
    Arun Kandoor\textsuperscript{2} &
    James Laudon\textsuperscript{2} \\
  \end{tabular} \\
  \begin{tabular}{cc}
    \textsuperscript{1}UC Berkeley &
    \textsuperscript{2}Google \\
  \end{tabular} \\
}

\begin{document}

\maketitle
\def\thefootnote{*}\footnotetext{Equal contribution.}
\def\thefootnote{\dag}\footnotetext{The work was partially completed
during Peter Schafhalter's internship at Google.}
\def\thefootnote{\ddag}\footnotetext{Corresponding author: Yanqi Zhou
<yanqiz@google.com>.}

\begin{abstract}
  Domain-specific adaptation is critical to maximizing the performance of
  pre-trained language models (PLMs) on one or multiple targeted tasks,
  especially under resource-constrained use cases, such as edge devices.
  However, existing methods often struggle to balance domain-specific
  performance, retention of general knowledge, and efficiency for training
  and inference.
  To address these challenges, we propose \modelnamefull (\modelname).
  \modelname is a mixture-of-experts architecture that
  augments a general PLMs with modular, domain-specialized experts.
  These experts are trained independently and composed together via a
  lightweight training process.
  In contrast to standard low-rank adaptation methods, each \modelname
  expert consists of several transformer layers which scale better with
  more training examples and larger parameter counts.
  Our evaluation demonstrates that \modelname achieves comparable target
  performances to full parameter fine-tuning while achieving 1.65\%
  better retention performance.
  Moreover, \modelname's architecture enables flexible sharding
  configurations and improves training speeds by up to 38\% over
  state-of-the-art distributed training configurations.
\end{abstract}

\section{Introduction}
\label{s:introduction}

Recent advances in large-scale Pre-trained Language Models (PLMs) have
showcased impressive generalization capabilities~\cite{brown2020language,
chowdhery2023palm,anil2023palm,team2023gemini}.
However, when applied to specialized domains such as medical, legal, or
financial sectors, these general-purpose models often require further
fine-tuning to maximize performance on target
domains~\cite{huang2023lawyer,li2023cfgpt,singhal2023large}. 

A straightforward approach to domain adaptation is full-parameter
fine-tuning, where the entire model is further trained on domain-specific
data~\cite{houlsby2019parameter, bapna2019simple}.
While this method provides strong performance on target domains,
full-parameter fine-tuning may lead to \textit{catastrophic forgetting}
where the model loses previously learned capabilities by overfitting to the
target domain~\cite{goodfellow2013empirical, kirkpatrick2017overcoming}.
Additionally, this method is memory-intensive to serve in multi-domain
settings as each domain has a unique set of parameters, incurring a
significant parameter loading overhead when switching between
domains~\cite{hu2021lora}.
In such cases, the cost of frequent ``context switches''
significantly impacts performance, making full-parameter fine-tuning
impractical for scalable, efficient deployment~\cite{dhar2024empirical}.

To address the issues of forgetting and memory-efficiency,
parameter-efficient fine-tuning methods have been proposed, such as adapter
modules~\cite{houlsby2019parameter}, LoRA~\cite{hu2021lora}, and
CoDA~\cite{lei2023conditional}.
These methods introduce a small number of trainable parameters and keep the
original model frozen during training.
Through targeted parameter updates,
these approaches are both computationally efficient and effective at
retaining prior knowledge~\cite{biderman2024lora}.
Despite these benefits, %
parameter-efficient methods are limited in their expressive potential and
struggle to scale effectively across large domains and
datasets~\cite{anyscale-finetuning-lora}.

In this work, we propose \modelnamefull (\modelname), a scalable method for
multi-domain adaptation. \modelname is inspired by the Mixture of Experts
(MoE) approach (detailed in \cref{s:background}), and extends PLMs by
introducing modular, composable domain-specialized experts.
In \modelname, each expert consists of several transformer layers that
operate in parallel to the ``backbone'' PLM.
During training, the backbone remains frozen, and
each expert is independently trained for its respective domain.
For multi-domain deployment, \modelname combines multiple experts with
different specializations by placing them in parallel to the backbone
PLM, allowing for strong performance across diverse domains.
A lightweight fine-tuning steps teaches the experts to collaborate
effectively to provide strong performance across all target domains.
We further take advantage of parallelism in \modelname's
architecture to enable new sharding strategies that improve training
efficiency, and provide privacy benefits during inference.

In summary, we make the following contributions:

\begin{itemize}
  \item We introduce the novel \modelnamefull (\modelname) architecture
    designed to address the scalability, catastrophic forgetting, and
    memory constraints in multi-domain adaptation.
  \item We evaluate the performance of \modelname and demonstrate that it
    outperforms LoRA by 1.4\% and full parameter fine-tuning by 0.6\% on a
    multi-domain task.
  \item We analyze the training efficiency of \modelname, showing better
    scalability compared to LoRA as the number of training examples and
    parameters increases.
  \item We demonstrate how \modelname's design enables flexible sharding
    strategies that accelerate training, achieving up to 38\% speedup over
    standard distributed training configurations.
\end{itemize}

\begin{figure} \centering
  \begin{subfigure}[b]{.62\textwidth}
    \centering
    \captionsetup{width=0.95\linewidth}
    \includegraphics[width=0.95\columnwidth]{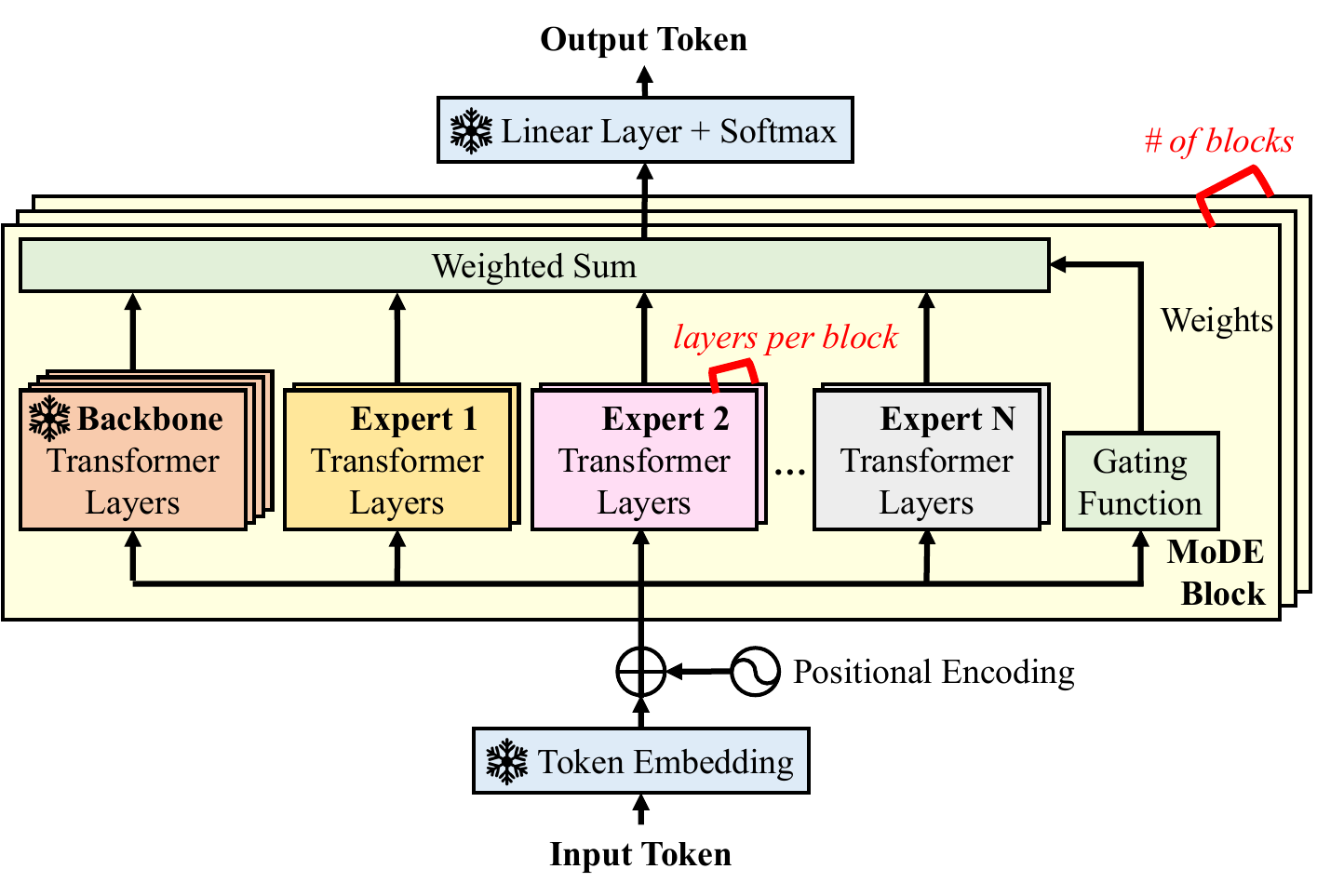}
    \caption{\small Model architecture.}
    \label{f:model-architecture}
  \end{subfigure}
  \begin{subfigure}[b]{.37\textwidth}
    \centering
    \captionsetup{width=0.95\linewidth}
    \includegraphics[width=0.95\columnwidth]{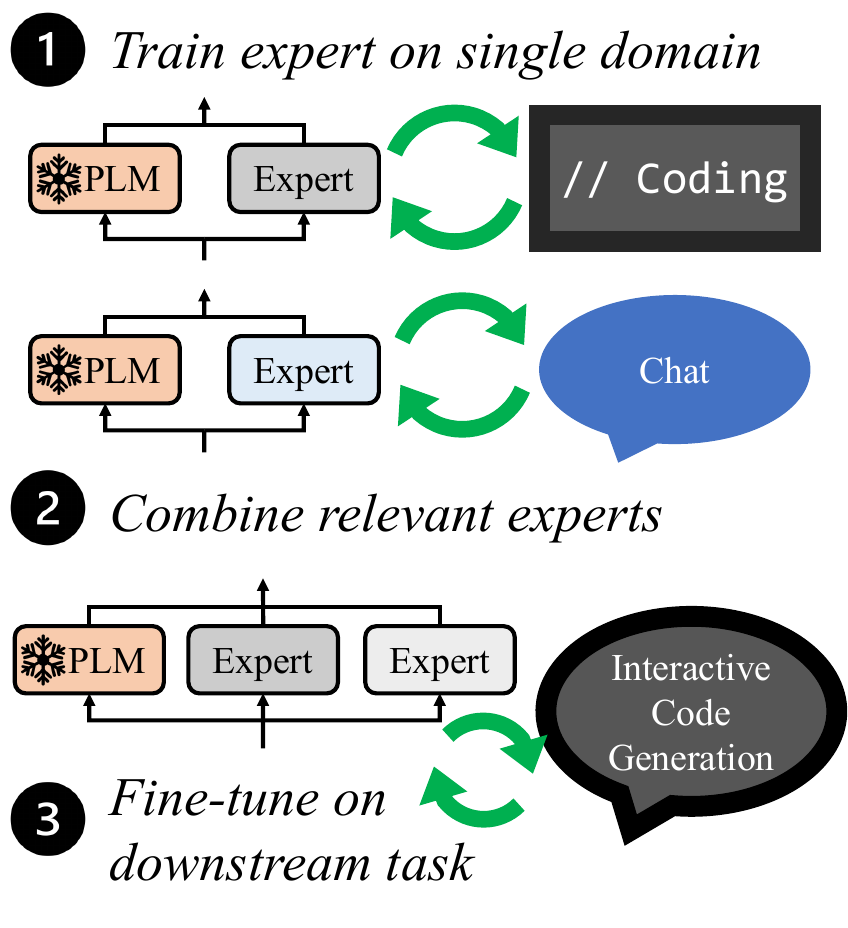}
      \caption{\small Training procedure.}
    \label{f:training-procedure}
  \end{subfigure}
  \caption{
    \small
    \textbf{\modelname overview.}
    \modelname models are divided into blocks, each containing transformer
    layers from the backbone or an expert (\cref{f:model-architecture}).
    Backbone and expert blocks operate on the same inputs.
    The model takes a linear combination of their outputs, where the weights
    are determined by a gating function.
    \cref{f:training-procedure} outlines the training process:
    \circled{1} Experts are trained independently on specific domains,
    while the backbone's parameters remain unchanged.
    \circled{2} For a multi-domain task, experts are modularly composed to
    enhance the model's performance. Here, the code and chat experts are
    combined to improve the performance of an interactive coding assistant.
    \circled{3} A lightweight fine-tuning process updates the experts and
    the gating function to improve performance on the target task.
    }
  \label{f:eot-method}
  \vspace{-1em}
\end{figure}

\section{Related Work}
\label{s:background}

\myparagraph{Domain-Adaptive Pre-training.}
\modelname is designed for scenarios where large domain-specific unlabeled
datasets are available, a process often referred to as ``domain-adaptive
pre-training'', ``continual learning'', ``continued pre-training'', or
``further pre-training''~\cite{shi2024continual, azerbayev2023llemma,
colombo2024saullm, agarwal2024structured}.
Beyond single-domain adaptation, recent research has expanded into
multi-domain adaptation, which presents additional
challenges~\cite{saunders2022domain, wu2024mixture}.
To the best of our knowledge, the most relevant works are
parameter-efficient fine-tuning methods, which are described in detail in
the next paragraph.

\myparagraph{Parameter-Efficient Fine-Tuning.}
Significant progress has been made in parameter-efficient fine-tuning
methods, which update only a small subset of model
parameters~\cite{houlsby2019parameter, he2021towards}.
Techniques such as LoRA~\cite{hu2021lora} and
QLoRA~\cite{dettmers2023qlora} enhance language model performance by
introducing trainable low-rank decomposition matrices.
However, these low-rank approaches often fall short when adapting to
domains that require high expressiveness, such as mathematical
reasoning or coding~\cite{biderman2024lora, anyscale-finetuning-lora}.

\myparagraph{Mixtures of Experts} (MoE) architectures are promising for
expanding the capacity of language models while keeping the computational
cost of training constant.
MoE models learn a routing function that selectively activates a subset of
experts, enabling sparse computation~\cite{du2022glam,
zhou2022mixtureofexperts}.
Unlike \modelname, MoE models are primarily used to enhance pre-training
performance~\cite{dai2024deepseekmoe, dbrx, grok1}.
Recent approaches extend MoE concepts to improve adaptation to new
domains, such as life-long learning with distribution-specialized
experts~\cite{chen2023lifelong}, conditional
adapters~\cite{lei2023conditional}, AdaMix~\cite{wang2022adamix}, and
MixPHM~\cite{jiang2023mixphm}.
These approaches modify PLMs at a fine granularity (\eg by modifying the
feed-forward network in the transformer layers).
In contrast, \modelname applies MoE at the transformer-level, offering a
scalable and expressive solution for multi-domain adaptation.

Furthermore, MoE-inspired approaches have shown potential in selecting
domain-specific adapters by routing input sequences, which enhances model
performance across diverse domains while alleviating catastrophic
forgetting~\cite{feng2024mixture}.
In this work, we focus on token-level routing, with sequence-level routing
as a natural extension for future research.

\myparagraph{Distributed Training.}
Scaling PLMs to billions of parameters requires partitioning parameters and
models and training data across multiple processors~\cite{dean2012large,
li2014scaling, barham2022pathways}.
Common sharding strategies, including \textit{model parallelism} and
\textit{data parallelism}, evenly divide inputs and parameters across
accelerators~\cite{zero,lepikhin2020gshard,alabed2024partir}.
To address memory overheads and communication bottlenecks, researchers have
developed specialized tools to optimize the sharding configuration of a
model for the available
processors~\cite{lepikhin2020gshard,alabed2024partir}.

Typically, distributed training and serving frameworks implement the 
\textit{Single Program, Multiple Data} (SPMD) model of computation which
enables building models and training programs as if developing for a single
processor with a large pool of memory.
However, SPMD has two key limitations:
\one{} all operations (i.e., layers of a model) execute sequentially,
precluding the execution of different operations in parallel on different
processors, and
\two{} all operations must use all devices, which may be inefficient (e.g.,
by incurring significant communication overhead).
In contrast, the \textit{Multiple Program, Multiple Data (MPMD)}
computational model has the ability to run multiple programs in parallel on
different devices, enabling fine-grained control over parallelism which has
the potential for more efficient execution compared to
SPMD~\cite{zheng2022alpa}.
We exploit parallelism in \modelname's structure to implement an
MPMD-enabled sharding strategy that accelerates training by up to 38\% over
SPMD model parallelism.

\section{Method}
\label{s:design}

Our proposed model architecture augments PLMs with modular, domain-specialized experts and aims to achieve the following design goals:
\begin{itemize}
  \item \textbf{Domain specialization.} Experts must provide strong
    performance and high expressiveness on their target domains.
    Moreover, composing multiple experts with a PLM should result in strong performance on each of the target domains.
  \item \textbf{Capability retention.} The model must preserve the
    generalist capabilities of the PLM and avoid catastrophic forgetting
    which may cause performance regressions on other tasks.
  \item \textbf{Efficient training and inference.} The model architecture
    must be efficient to train on large clusters, and provide advantages
    for inference on edge devices.
\end{itemize}

\subsection{Model Design}
\label{s:design:model-design}

To meet these design goals, we decompose the language model into multiple
\modelname \textit{blocks} (see \cref{f:model-architecture}). Each $i$th
\textit{block} contains:
\one{} a few backbone transformer layers,
$f_{\text{bb}}^i(\cdot)$, which are initialized from PLMs, and
\two{} a set of $N$ experts where each expert has the same (small) number
of transformer layers, and the $i$th block of the $j$th expert is given by 
$f_{\text{e}}^{i, j}(\cdot)$.
All backbone and expert layers run in parallel, with their outputs combined
as a weighted sum.
The weights for each component are determined by a lightweight gating
function, $g^i(\cdot)$.

For each $i$-th block, the input $x \in \mathbb{R}^d$ and the output $y \in \mathbb{R}^d$ are defined as:
\begin{align}
  y &= \alpha_{\text{bb}} \, f_{\text{bb}}^i(x) + \sum_{j=1}^N
  \alpha_{\text{e}}^j \, f_{\text{e}}^{i,j}(x) \\
  &\quad \text{where} \quad [\alpha_{\text{bb}}, \alpha_{\text{e}}^1,
  \alpha_{\text{e}}^2, \cdots, \alpha_{\text{e}}^N] = g^i(x) \nonumber
\end{align}

\myparagraph{Backbone Model.}
The backbone transformer layers, $f_{\text{bb}}^i(x)$ are derived from
PLMs. In each \modelname block, the number of backbone layers is determined
by dividing the total number of layers in the original PLM by the total
number of \modelname blocks.
The number of \modelname blocks is an important configuration
hyperparameter:
more
blocks enables more frequent
synchronization between the backbone and expert layers which provides
greater flexibility.
In \cref{t:moa-config-ablation}, we conduct an ablation to assess the
impact of the number of blocks on accuracy.
In \modelname, the backbone layers are initialized from PLMs and remain
frozen during training to mitigate the issue of catastrophic forgetting.

\myparagraph{Experts.}
Each expert, $f_{\text{e}}^{i,j}(x)$ consists of several transformer
layers, and the number of transformer layers impacts the expert's
performance.
\cref{t:moa-config-ablation} indicates that increasing the number of expert
layers improves performance on the target domain at the cost of higher
computational cost.
We use transformer layers for experts for two reasons:
\one{} transformer blocks scale more efficiently than finer-grained designs,
such as modifications within attention matrices, and
\two{} they simplify deployment, as there are no modifications to the
transformer architecture itself.
In contrast, fine-grained methods often face deployment
challenges~\cite{yi2023edgemoe}.

\myparagraph{Gating Function.}
The outputs from the backbone and expert layers are combined through the gating layer, $g^i(x)$.
In this work, we use a simple token-level gating function.
This function consists of linear layer followed by a softmax for
normalization.
The gating function takes $x$ as input and outputs a vector of length $N+1$
as follows:
\begin{align}
  g^{i}(x) = \texttt{softmax}(\texttt{Linear}(x))
\end{align}

The simplicity of \modelname's gating function enables efficient
composition of multiple experts, as shown in \cref{t:accuracy}.
However, future work could explore more advanced designs, such as sparse
gating and sequence routing, to further optimize memory usage and
computational efficiency.

\begin{figure}
  \centering
  \begin{subfigure}[b]{0.49\textwidth}
    \centering
    \captionsetup{width=0.95\linewidth}
    \includegraphics[width=0.9\columnwidth]{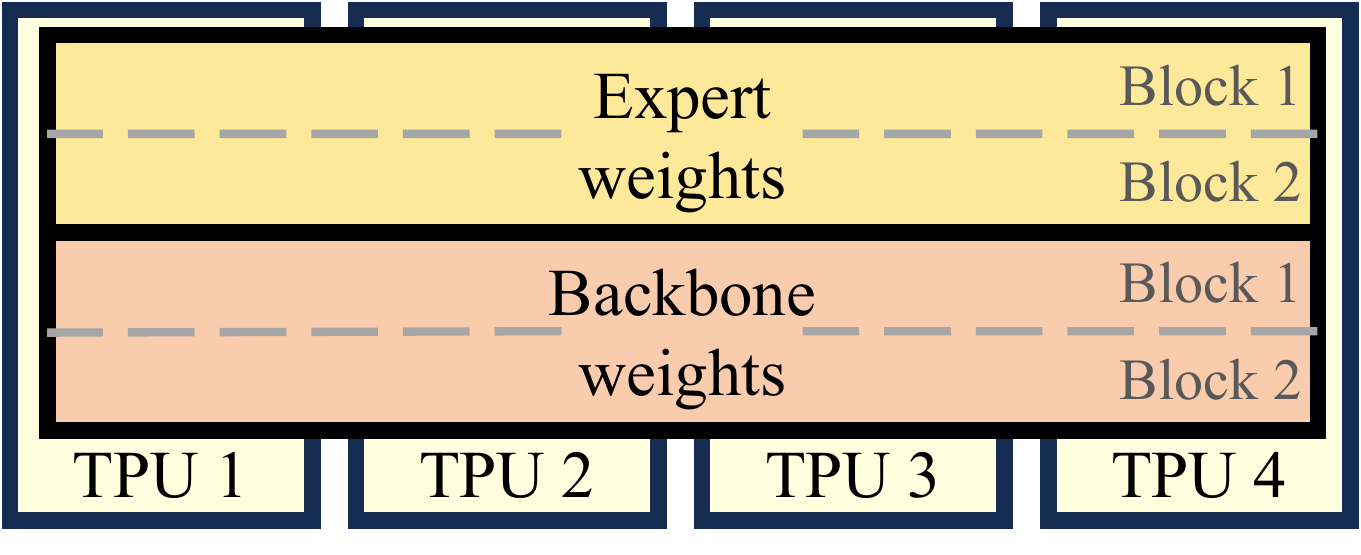}
    \caption{\small A model-parallel partitioning in which an entire
    \modelname model is split across all TPUs.}
    \label{f:model-parallel-sharding}
  \end{subfigure}
  \begin{subfigure}[b]{0.49\textwidth}
    \centering
    \captionsetup{width=0.95\linewidth}
    \includegraphics[width=0.9\columnwidth]{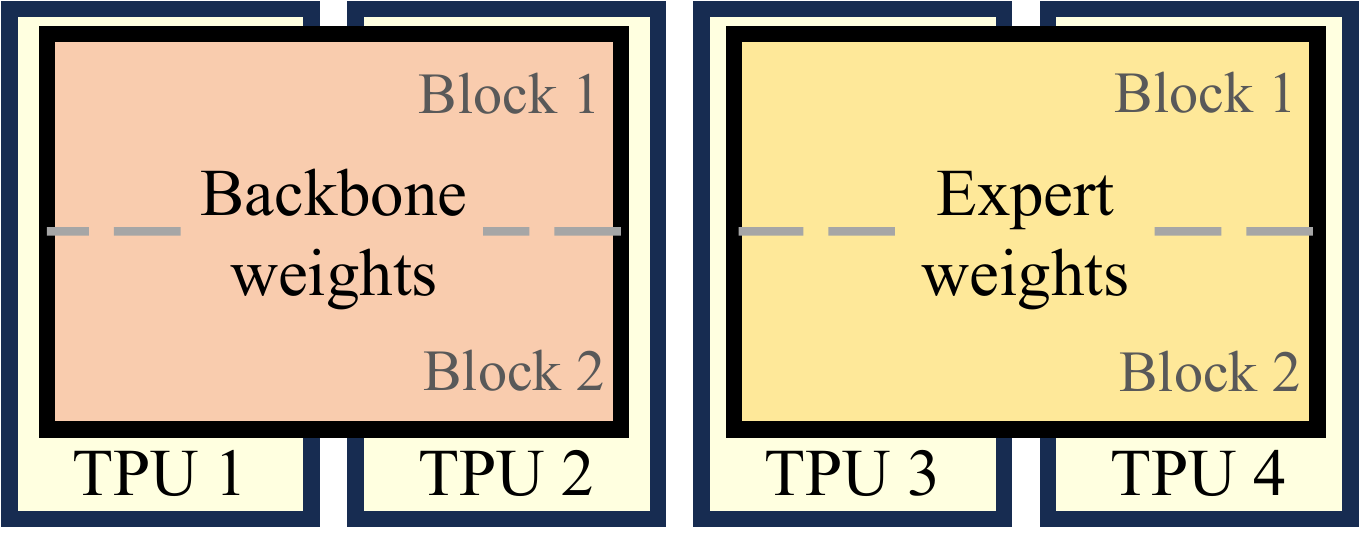}
    \caption{\small An flexible sharding configuration enabled by the \modelname
    architecture and MPMD.
    }
  \end{subfigure}
  \caption{\small \textbf{Example SPMD and MPMD sharding configurations.}
  While the model parallel sharding configuration enabled by SPMD evenly
  distributes all weights across all TPUs, the provided MPMD sharding
  configuration executes the backbone and the expert on 2 different
  meshes consisting of 2 TPUs each which may reduce communication
  overheads.
  }
  \label{f:eot-mpmd-sharding}
\end{figure}

\subsection{Training Procedure}
\label{s:design:training-procedure}
\modelname has a two stage training procedure (\cref{f:training-procedure}): $(i)$ train a single expert independently on each domain, and $(ii)$ compose different experts into a single model to improve multi-domain performance.

\myparagraph{Single Modular Expert Training.}
For each domain, we independently train an \modelname model with a single expert. During training, only the expert layers and gating layers are updated, while the backbone layers remain frozen. Freezing the backbone not only reduces the computational cost of backpropagation but also helps mitigate catastrophic forgetting.

\myparagraph{Composing Experts.}
After training the modular experts on each domain, we can compose them into a single \modelname model for multi-domain adaptation. While the experts contain domain-specific knowledge, their outputs can interfere when combined, as they are pre-trained individually. To address this, we apply a lightweight fine-tuning step using data from all domains to learn the optimal weights for the gating function. Although freezing the experts requires fewer training examples during composition \cref{s:mixture-data-efficiency}, we have demonstrated that unfreezing the experts during this step further enhances multi-domain adaptation performance \cref{t:accuracy}.

\subsection{Parallelizing Block Execution}
\label{s:design:adapter-execution}

We exploit parallelism in \modelname's structure to enable flexible
sharding configurations supported by the MPMD model of computation.
These sharding configurations improve training efficiency due to:
\begin{enumerate}
  \item The ability to execute the backbone and experts on distinct,
    smaller sets of devices which reduces communication overheads
    (\cref{f:eot-mpmd-sharding}).
  \item The ability to configure the backbone and experts independently to
    maximize overall efficiency, when their number of parameters are
    different. 
\end{enumerate}

While MPMD has the potential to reduce communication overheads by running
blocks on fewer devices, the merges in the \modelname model architecture
are points of synchronization that require communication across all
devices.
To determine whether MPMD is beneficial for training, we evaluate whether
MPMD configurations can afford the cost of resharding between meshes in
\cref{s:evaluation:mpmd-training}.

While MPMD has the potential to reduce communication overheads by executing
blocks on smaller sets of devices, the linear combination of the outputs
from the backbone and expert blocks requires are points of synchronization
that require communication across all devices.
To assess the effectiveness of MPMD in training, we evaluate if the
benefits of MPMD configurations outweigh the costs associated with
resharding between device meshes, as discussed in
\cref{s:evaluation:mpmd-training}.

Moreover, \modelname's architecture enables several benefits during
inference for privacy and performance.
By using MPMD to run \modelname with the backbone and expert parameters on
different devices, expert parameters for sensitive domains (\eg personal
information, medical data, etc.) can remain private (\eg by executing on a
user's phone).
To switch to another \modelname model adapted to different task, only the
new expert weights need to be loaded into memory because the backbone PLM's
weights are shared. %
This reduces the overhead of loading model weights and works well with
existing techniques developer to serve many adapters concurrently, such as
SLoRA~\cite{sheng2024slora}.

\section{Experiments}
\label{s:evaluation}

\subsection{Experiment Setup}
\label{s:eval:setup}

\myparagraph{PLM Configuration.}
Our PLM model consists of 1.58 billion parameters distributed across 18 transformer layers, comparable to smaller open-source models such as Gemma~\cite{team2024gemma}, Phi~\cite{gunasekar2023textbooks}, and Llama 3.2~\cite{llama-3.2}. The model is pre-trained on a high-quality dataset that spans a diverse range of natural language use cases, similar to GPT-3~\cite{brown2020language}, GLaM~\cite{du2022glam}, and LLaMA~\cite{touvron2023llama}.

\myparagraph{Multi-Domain Datasets.}
We prepare two target domain datasets, \textit{Code} and \textit{Math}, to evaluate multi-domain adaptation, and one retention dataset, \textit{English}, to measure catastrophic forgetting:

\begin{itemize} \item \textit{Code} consists of code samples retrieved from real-world applications. \item \textit{Math} tests the model’s mathematical reasoning capabilities and is similar to the GSM~\cite{cobbe2021training} and MATH~\cite{hendrycks2021measuring} datasets. \item \textit{English} contains literature texts with a distribution distinct from \textit{Code} and \textit{Math}. \end{itemize}

Additionally, we create a mixed dataset, \textit{Math + Code}, for model training, which contains an equal number of samples from the \textit{Math} and \textit{Code} datasets.

\myparagraph{Evaluation Metric.}
As our focus is on domain-adaptive pre-training with large amounts of unlabeled data, we adopt next-token prediction accuracy as the primary metric to assess the effectiveness of our methods. This metric has been shown to correlate strongly with downstream performance~\cite{kaplan2020scaling, hoffmann2022training}. In addition, since the evaluation spans multiple domains and includes a retention dataset, we use the average performance across all datasets to rank the models.

\myparagraph{Baselines.}
We benchmark \modelname against two widely-used domain adaptation methods: full-parameter fine-tuning~\cite{houlsby2019parameter} and LoRA~\cite{hu2021lora}.
We exclude similar low-rank methods, such as CoDA~\cite{lei2023conditional} and QLoRA~\cite{dettmers2023qlora}, which prioritize computational efficiency and exhibit lower accuracy than LoRA.
Additionally, we conduct a comprehensive hyperparameter search for LoRA (\cref{t:lora-config-ablation}) and report results for the best-performing LoRA configuration.

\myparagraph{Chosen \modelname Configuration}. We configure \modelname with three blocks, each containing two transformer layers per expert. For a detailed comparison of different configurations, refer to \cref{t:moa-config-ablation}.

\myparagraph{Training Details} All methods are trained on a cluster of 128 TPUv3 accelerators across 8 servers. We use a batch size of 128, a learning rate of 0.001,
and train for 50k steps. %

\subsection{Multi-Domain Adaptation}
\label{s:evaluation:multi-domain-adaptation}

\myparagraph{Overall Performance.} 
We evaluate \modelname, LoRA, and full-parameter fine-tuning on the \textit{Math}, \textit{Code}, and \textit{English} test sets.
We train the baseline models
directly on the mixed dataset, \textit{Math + Code}.
For \modelname, we follow the training procedure described in \cref{s:design:training-procedure}, where individual \modelname models are first trained on \textit{Math} and \textit{Code} training sets independently, and then composed using the \textit{Math + Code} dataset.
As shown in \cref{t:accuracy}, \modelname with two experts (one for \textit{Math} and one for \textit{Code}) achieves the best overall performance and retention capabilities. On average, \modelname outperforms full-parameter fine-tuning by 0.59\% and LoRA by 1.41\%.
Notably, \modelname achieves target domain performance comparable to full-parameter fine-tuning, outperforming it by 0.28\% on \textit{Math} while trailing by 0.06\% on \textit{Code}, and provides a 1.55\% improvement in retention on the \textit{English} dataset.
Compared to LoRA, \modelname delivers a higher accuracy on the target domains, with a 1.68\% improvement on \textit{Math} and 2.12\% on \textit{Code}, and surpasses LoRA by 0.45\% in \textit{English} retention.

\begin{table}[!t]
  \centering
  \begin{tabular}{lc|cccc}
\toprule
  \multicolumn{2}{c|}{\textbf{Method}} &
  \multicolumn{4}{c}{\textbf{Accuracy}} \\
  {Method} & {Parameters} & {Math} & {Code} & {English} & {Average} \\
  \midrule
  Full-parameter Fine-tuning & 1.583 B & 77.18 & \textbf{67.89} & 47.95 &
  64.34 \\
  LoRA & 1.585 B & 75.79 & 65.71 & 49.05 & 63.52 \\
  \midrule
  \modelname 1$\times$Uninitialized Expert & 1.979 B & 76.71 & 67.30 & 49.17 & 64.39 \\
  \modelname 2$\times$Uninitialized Experts & 2.376 B & 76.87 & 67.39 &
  49.47 & 64.58 \\  
  \modelname 2$\times$Frozen Experts & 2.376 B & 76.99 & 66.94 & 49.07 &
  64.33 \\  \midrule
  \modelname 2$\times$Experts & 2.376 B & \textbf{77.47} & 67.83 &
  \textbf{49.50} & \textbf{64.93} \\
  $\quad$ vs. Full-parameter
  Fine-tuning & $\uparrow$0.793 B & $\uparrow$0.29 & $\downarrow$0.06 &
  $\uparrow$1.65 & $\uparrow$0.59 \\
  $\quad$ vs. LoRA & $\uparrow$0.789 B & $\uparrow$1.68
  & $\uparrow$2.12 & $\uparrow$0.45 & $\uparrow$1.41 \\
  \bottomrule
\end{tabular}

  \caption{Multi-domain performance comparison of several methods which
  demonstrates that \modelname provides the best overall performance and
  capability retention.
  We adapt each method to the \textit{Math + Coding} dataset, and compare
  the performance on the test sets for math, coding, and English retention.
  }
  \label{t:accuracy}
\end{table}

\myparagraph{\modelname Configurations.}
We address two questions in exploring different configurations: 
\begin{enumerate} \item Which configuration of \modelname modular experts provides the best performance? \item Which composition strategy yields the best multi-domain performance? \end{enumerate}

For the first question, we evaluate \modelname with only one uninitialized expert and vary two hyperparameters: the number of \modelname blocks, and the number of transformer layers per expert within each block. As shown in \cref{t:moa-config-ablation}, increasing the total number of expert layers leads to higher accuracy in the target domain.
To balance the number of added parameters, domain-specific accuracy, and capability retention, we configure the \modelname model with three \modelname blocks, each containing two transformer layers per expert (six layers per expert total).

For the second question, we compare against three alternative \modelname configurations.
We first measure the benefit of re-using experts by training \modelname configurations with one or two randomly initialized experts:
\modelname 1$\times$Uninitialized Expert measures the techniques performance when training a new expert on the multi-domain dataset from scratch, and \modelname 2$\times$Uninitialized Experts accounts for the difference in the number of parameters.
We also measure how much fine-tuning on the multi-domain dataset benefits accuracy.
For this configuration (\modelname 2$\times$Frozen Experts), we train only the gating function and re-use the weights single-domain training: one expert trained on \textit{Math} and the other trained on \textit{Code}.
According to \cref{t:accuracy}, we find that more experts, initialization, and multi-domain fine-tuning all improve performance.

In addition, we examine how much mixed data from \textit{Math + Code} is required for strong performance. %
As discussed in \cref{s:mixture-data-efficiency}, freezing the experts during composition yields better training sample efficiency compared to other methods.
This finding highlights the potential of our approach for privacy-preserving training, where different entities can independently train domain-specialized experts and later combine them into a coherent model, without sharing expert weights and only sharing sharing a small amount of training data.

\begin{figure*}
  \centering
  \begin{subfigure}[t]{.42\textwidth}
    \centering
    \captionsetup{width=0.95\linewidth}
    \includegraphics[width=0.95\columnwidth]{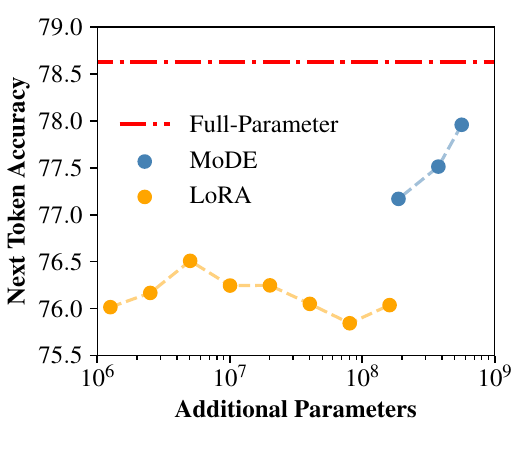}
    \label{f:parameter-scalability}
  \end{subfigure}
  \begin{subfigure}[t]{.42\textwidth}
    \centering
    \captionsetup{width=0.95\linewidth}
    \includegraphics[width=0.95\columnwidth]{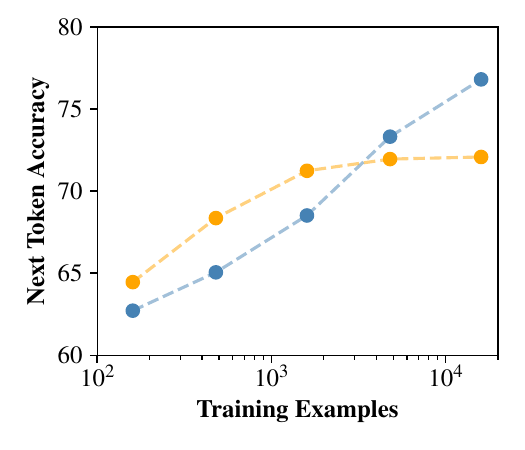}
    \label{f:training-example-scalability}
  \end{subfigure}
  \caption{
    \small
    \textbf{Scalability of adaptation methods.}
    We find that \modelname scales better than LoRA as the number of training examples and the number of parameters added increases. In the left figure, we increase the number of adapter parameters for LoRA by increasing the rank and \modelname by increasing the number of expert layers, and find that \modelname provides higher accuracy than LoRA with more trainable parameters.
    In the right figure, we generate versions of the \textit{Code} with different numbers of training examples, and train a LoRA adapter and a \modelname expert for the same number of training steps on each.
    Although LoRA provides better accuracy on small datasets up to $\sim$1k training examples,
    we find that \modelname's accuracy is better on large datasets, demonstrating that \modelname scales better with more training data than LoRA.
}
\label{f:scalability-analysis}
\end{figure*}

\subsection{Scalability of \modelname}
\label{s:eval:modular-training}

We evaluate the scalability of \modelname along two dimensions: scaling with additional parameters and scaling with increased training examples. We use LoRA as the baseline in both experiments and conduct experiments on a single domain using the \textit{Code} dataset for training and evaluation.

\myparagraph{Scalability with Additional Parameters.}
We compare the accuracy improvements of \modelname and LoRA as additional parameters increases. For LoRA, we increase parameters by increasing the rank of the decomposition matrices, ranging from 8 to 2,048. For \modelname, we increase the number of transformer layers per expert while keeping the number of \modelname blocks constant. As shown in \cref{f:scalability-analysis}, while LoRA is parameter-efficient, it struggles to convert additional parameters into higher accuracy. In contrast, \modelname shows a continuous improvement in accuracy as more parameters are added, indicating that \modelname scales more effectively with more parameters.

\myparagraph{Scalability with Training Examples.}
We evaluate how well \modelname and LoRA leverage additional training data to improve accuracy. Different versions of the \textit{Code} dataset were created, each containing varying amounts of training examples. Both \modelname and LoRA were trained on these dataset versions, with the number of training steps keep the same across all experiments. As shown in \cref{f:scalability-analysis}, we observe that LoRA performs better on smaller datasets (hundreds to thousands of training examples), but its accuracy plateaus as the number of training examples exceeds one thousand. In contrast, \modelname continues to improve as the number of training examples increases, demonstrating superior scalability with larger training sets.

\begin{figure*}
  \begin{subfigure}[t]{.34\textwidth}
\centering
\captionsetup{width=0.95\linewidth}
\includegraphics[width=0.95\columnwidth]{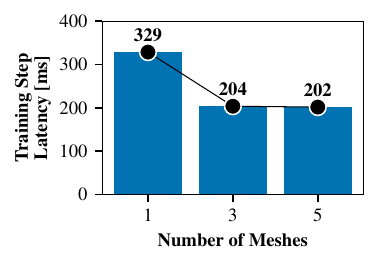}
\caption{\small The cost of resharding between meshes is lower than the
    cost of model parallelism with experts.}
\label{f:cost-of-resharding}
\end{subfigure}
  \begin{subfigure}[t]{.30\textwidth}
\centering
\captionsetup{width=0.95\linewidth}
\includegraphics[width=0.95\columnwidth]{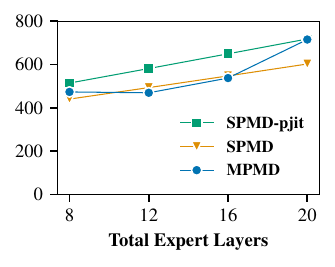}
\caption{\small MPMD enables faster training when the latency from the
    expert layers is close to the latency from the backbone.}
\label{f:cost-of-layers-added}
\end{subfigure}
  \begin{subfigure}[t]{.31\textwidth}
\centering
\captionsetup{width=0.95\linewidth}
\includegraphics[width=0.95\columnwidth]{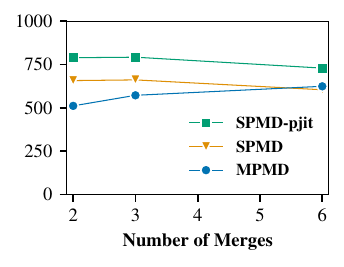}
\caption{\small As the amount of merges increases, the speed of MPMD
    training decreases due increased communication costs.}
\label{f:cost-of-merges}
\end{subfigure}
  \caption{\small \textbf{Evaluation of flexible sharding configurations}
  enabled by the \modelname model architecture.}
 \vspace{-1em}
\end{figure*}

\subsection{Accelerating Training}
\label{s:evaluation:mpmd-training}

Through our experiments, we find several MPMD-enabled
sharding configurations which increase training
speeds by up to 38\% over SPMD model-parallelism.
The results confirm that the reduction in communication costs by running
backbones and experts on fewer accelerators outweighs the added costs of
resharding across meshes when merging the intermediate outputs of models.

All experiments use 8 TPUv5e accelerators which are managed by the Pathways
orchestration layer which supports MPMD~\cite{barham2022pathways}.
We implement MPMD training by extending the Pax distributed machine learning
framework 
which is
implemented in Jax~\cite{jax2018github}. While we use TPUs and a Jax-based framework as our training environment, we
underscore that the key idea of using MPMD to increase training speed in distributed settings extends to other accelerators (e.g., GPUs) and
training systems such as PyTorch~\cite{paszke2019pytorch}. We measure the training performance, but note that many of the performance improvements enabled by MPMD sharding strategies may also benefit model serving. In details, we compare the following configurations:
\begin{enumerate}
  \itemsep0em
  \item \textbf{MPMD} divides the TPUs
    into different meshes to which backbone or experts are assigned.
  \item \textbf{SPMD} creates a single
    mesh which consists of all 8 TPUs, backbone and experts.
  \item \textbf{SPMD-pjit} uses our training library's to parallelize model training.
\end{enumerate}

\myparagraph{Cost of Resharding.}
We compare the following configurations, which train \modelname using 4
experts where each consists of 6 transformer layers divided evenly into two
blocks:
\begin{enumerate}
  \itemsep0em
  \item \textit{1 mesh} uses model parallelism to shard the model across a
    single mesh.
  \item \textit{3 mesh} assigns the backbone to 4 TPUs and each expert
    shares 2 TPUs with another expert. The backbone and the experts
    are sharded on their meshes using model parallelism.
  \item \textit{5 mesh} assigns the backbone to 4 TPUs, and one expert to
    each of the remaining TPUs. The backbone is sharded with model
    parallelism.
\end{enumerate}
We find that the cost of resharding between meshes is lower than the cost
of model parallelism (\cref{f:cost-of-resharding}).
The reduction in training step latency from $329$ ms for \textit{1 mesh} to $204$
ms, a 38\% reduction, results due to the reduction in communication caused
by the MPMD sharding configuration.
Because the latency of the \textit{3 mesh} and \textit{5 mesh}
configurations is similar, we conclude that the cost to reshard does not
increase with the number of meshes.

\myparagraph{Configuring the Expert Size.}
We configure a mixture of 2 expert divided into two blocks and change the
total number of transformer layers added. 
For example, setting the expert block size to 2 transformer layers adds
8 total transformer layers to the model.
The backbone is assigned to 6 TPUs with a model parallel sharding, and each
expert is assigned to one unoccupied TPU.

We find that SPMD performs better than MPMD when the expert size is small
or large (\cref{f:cost-of-layers-added}) due to \textit{under-utilization}
of the TPUs.
When the expert size is small (e.g., 8 added layers total) and we train
using MPMD, the expert blocks finish executing before the backbone block,
so the expert TPUs remain idle until the backbone block completes and the
merge begins.
Similarly, when the expert size is large (e.g., 20 added layers total),
the backbone blocks finish before the expert blocks so the backbone's TPUs
idle until the expert complete.
In contrast, SPMD executes the backbone and expert blocks sequentially on
all TPUs, ensuring that each TPU is always utilized.

When the runtime of a backbone block is similar to the runtime of the
expert blocks, MPMD training results in high utilization of the
accelerator.
Several settings impact the TPU utilization using MPMD which can be
configured to ensure high utilization, such as the number of accelerators
allocated to each mesh, the size of the expert blocks, and the sharding
strategy on each mesh.

\myparagraph{Cost of Merges.}
We train a mixture of 4 experts with 6 transformer layers each.
The backbone is assigned to a mesh of 4 TPUs, and each expert is assigned
to 1 unoccupied TPU.
We change the number of merges which determines the number of blocks (e.g.,
3 merges divide the backbone and each expert into three blocks).
We find that MPMD training speed decreases as the number of merges
increases due to the added communication costs incurred by each merge
(\cref{f:cost-of-merges}).

\section{Conclusions}
\label{conclusion}

Adapting pre-trained language models (PLMs) to complex, multi-domain
settings is increasingly important as these models are deployed in
specialized environments requiring diverse capabilities.
To address this challenge, we propose \modelnamefull (\modelname), a
scalable technique for adapting PLMs to multi-domain tasks.
\modelname introduces modular, domain-specialized experts while preserving
the general knowledge of the PLM by keeping its weights frozen.
\modelname allows experts to scale with both the number of parameters and
the amount of training data, outperforming standard parameter-efficient
methods like LoRA by 1.4\% on a challenging multi-domain dataset.
Additionally, \modelname delivers competitive performance compared to
full-parameter fine-tuning, achieving 1.5\% better retention of general
capabilities and 0.6\% higher accuracy across all subdomains.
To further optimize training efficiency, \modelname’s architecture supports
flexible sharding strategies through MPMD, resulting in up to 38\% faster
training compared to standard distributed training methods.
The ability to compose and reuse modular experts represents a significant
advancement in adapting PLMs to complex domains.
We hope that our work inspires further research into building modular,
expert-based language models.

\myparagraph{Acknowledgments.}
We would like to thank Ionel Gog and the Pathways Team for their support
and feedback on this work.

\bibliographystyle{iclr2025_conference}
\IfFileExists{references.bib}{
  \bibliography{references.bib}
}{
  \bibliography{../references.bib}
}

\newpage
\appendix
\section{Ablation of LoRA Configurations}
\label{s:lora-ablation}

In \cref{t:lora-config-ablation}, we provide an ablation study of
different LoRA configurations when adapting to the \textit{Code}.
We vary the rank of the low-rank matrices introduced for adaptation from 8
to 2048.
While a rank of 32 provides the best performance on the \textit{Code}
dataset, we find that a LoRA rank of 16 provides the best balance between
adaptation to \textit{Code} and capability retention on the
\textit{English} dataset.
We also test whether to apply LoRA to the feed-forward networks (FFN)
and the token embeddings in addition to the attention matrices.
We find that applying LoRA to the FFN slightly improves accuracy on
\textit{Code} by 0.27\% at the cost of a 0.86\% decrease in accuracy on
\textit{English} retention.

\begin{table}[h]
  \centering
  \begin{tabular}{ccc|ccc}
  \toprule
  \multicolumn{3}{c|}{\textbf{Configuration}} &
  \multicolumn{3}{c}{\textbf{Accuracy}} \\
  Rank & FFN & Embeddings & Code & English & Average \\
  \midrule
  8 & N & N & 76.02 & \textbf{48.76} & 62.39 \\
  \underline{16} & \underline{N} & \underline{N} & \underline{76.17} &
  \underline{48.64} & \underline{\textbf{62.40}} \\
  32 & N & N & \textbf{76.51} & 46.99 & 61.75 \\
  64 & N & N & 76.25 & 48.25 & 62.25 \\
  128 & N & N & 76.25 & 47.75 & 62 \\
  256 & N & N & 76.05 & 46.62 & 61.34 \\
  512 & N & N & 75.84 & 44.39 & 60.12 \\
  1024 & N & N & 76.04 & 43.91 & 59.97 \\
  2048 & N & N & 66.55 & 32.27 & 49.41 \\
  \midrule
  16 & Y & Y & 76.23 & 47.04 & 61.64 \\
  16 & Y & N & 76.44 & 47.78 & 62.11 \\
  16 & N & Y & 75.81 & 48.03 & 61.92 \\
  \underline{16} & \underline{N} & \underline{N} & \underline{76.17} &
  \underline{48.64} & \underline{\textbf{62.40}} \\
  \bottomrule
\end{tabular}

  \caption{Ablation of LoRA configurations when adapting  to the
  \textit{Code} dataset.}
  \label{t:lora-config-ablation}
\end{table}

\section{Ablation of \modelname Configurations}

We investigate how \modelname hyperparameters impact accuracy.
In \cref{t:moa-config-ablation}, we vary the number of blocks in a \modelname model and the number of expert transformer layers per block.
When adapting to the \textit{Code} dataset, we find that the number of blocks impacts both domain-specific and retention performance, indicating that tuning the number of blocks impacts model performance.
We further find that performance on \textit{Code} increases with the number of expert transformer layers which corresponds to the number of added parameters, but does not seem to significantly impact \textit{English} retention.

\begin{table}[h]
  \centering
  \begin{tabular}{cc|ccc}
\toprule
  \multicolumn{2}{c|}{\textbf{Model Configuration}} &
\multicolumn{3}{c}{\textbf{Accuracy}} \\
Blocks & Expert Layers & Code & English Retention & Average \\
\midrule
2 & 1 & 76.98 & 47.87 & 62.43 \\
2 & 2 & 77.17 & 47.86 & 62.51 \\
3 & 1 & 77.26 & 47.49 & 62.38 \\
  \underline{3} & \underline{2} & \underline{77.51} &
  \underline{\textbf{47.95}} & \underline{62.73} \\
6 & 1 & 77.79 & 47.61 & 62.70 \\
6 & 2 & \textbf{77.96} & 47.65 & \textbf{62.80} \\
\bottomrule
\end{tabular}

  \caption{Ablation of \modelname model configurations when fine-tuning on
  a coding dataset.
  We find that accuracy on the coding dataset increases with the number of
  expert layers.
  We select the underlined configuration with 3 blocks and a coding expert
  with 6 transformer layers to strike a balance between domain-specific
  accuracy, capability retention, %
  and the number of parameters added.}
  \label{t:moa-config-ablation}
\end{table}

\newpage

\section{Multi-Domain Data Efficiency}
\label{s:mixture-data-efficiency}

We explore how much multi-domain \textit{Math + Code} data is required for strong performance on the target \textit{Math} and \textit{Code} datasets as well as \textit{English} retention.
We follow a similar experimental setup to \cref{s:eval:modular-training}, but adapt the model to the multi-domain \textit{Math + Code} dataset instead of the single-domain \textit{Code} dataset.
A smaller amount of multi-domain data is desirable,
as it allows each domain owner to contribute minimal data while ensuring that their experts can collaborate effectively with those from other domains.

\begin{figure}[h]
  \centering
  \includegraphics[width=0.95\textwidth]{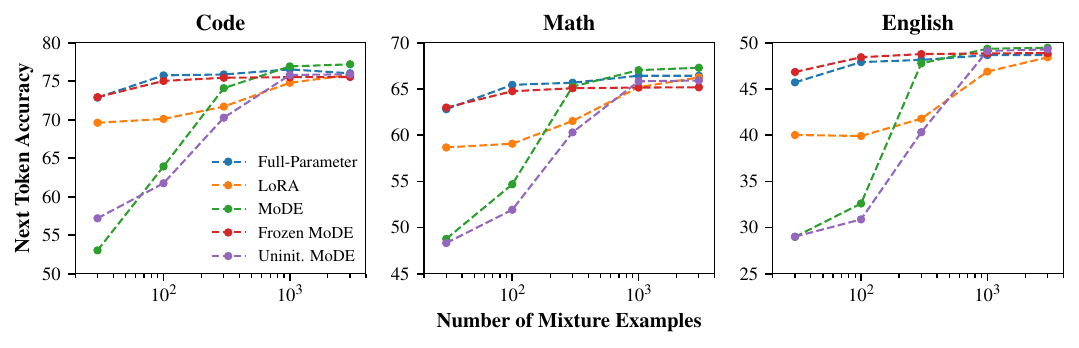}
  \caption{
    \small
    \textbf{Efficiency of mixture data.} We examine how much mixture data is required for training. For left to right, we present the next token accuracy on \textit{Code}, \textit{Math}, and \textit{English}. The x-axis for each subplot is the number of examples from \textit{Math + Code} mixture data. 
  \label{f:mixture_data}
}
\end{figure}

We compare against two \modelname variations alongside the baselines.
The first variation, Frozen \modelname, freezes experts during composition, while the second, Uninitialized \modelname, skips single-domain training entirely.
As shown in \cref{f:mixture_data}, Frozen \modelname delivers the best performance when less mixture data is available. However, with more training data, the standard \modelname configuration achieves the highest accuracy across all domains.

\end{document}